\documentclass[10pt, a4paper]{article}
\usepackage{pgfplots}
\pgfplotsset{compat=1.18}
\usepackage{pgfplotstable}
\usepackage{listings}
\usepackage{upquote}
\usepackage{xcolor}
\usepackage{graphicx}
\setkeys{Gin}{draft=false}
\usepackage{amsmath}
\usepackage{amssymb}

\lstdefinestyle{prompt}{
  basicstyle=\ttfamily\small,
  breaklines=true,
  breakatwhitespace=false,
  breakautoindent=false,
  breakindent=0pt,
  columns=fullflexible,
  keepspaces=true,
  showstringspaces=false,
  tabsize=2,
  xleftmargin=1em, xrightmargin=1em,
  frame=single, rulecolor=\color{black!25},
  backgroundcolor=\color{black!3},
  aboveskip=6pt, belowskip=6pt, captionpos=b
}

\lstdefinestyle{mdpretty}{
  basicstyle=\ttfamily\small,
  breaklines=true,
  breakatwhitespace=false,
  breakautoindent=false, 
  breakindent=0pt,
  columns=fullflexible,
  keepspaces=true,
  showstringspaces=false,
  tabsize=2,
  xleftmargin=1em, xrightmargin=1em,
  aboveskip=6pt, belowskip=6pt,
  frame=single, framerule=0.4pt,
  rulecolor=\color{black!25},
  backgroundcolor=\color{black!3}, 
  captionpos=t 
}

\usepackage[final]{lrec2026} 

\title{Adaptive Chunking: Optimizing Chunking-Method Selection for RAG}

\name{Paulo Roberto de Moura Júnior, Jean Lelong, Annabelle Blangero}

\address{Ekimetrics, France \\
         jean.lelong@ekimetrics.com, annabelle.blangero@ekimetrics.com\\}

\abstract{
The effectiveness of Retrieval-Augmented Generation (RAG) is highly dependent on how documents are chunked, that is, segmented into smaller units for indexing and retrieval. Yet, commonly used ``one-size-fits-all" approaches often fail to capture the nuanced structure and semantics of diverse texts. Despite its central role, chunking lacks a dedicated evaluation framework, making it difficult to assess and compare strategies independently of downstream performance. We challenge this paradigm by introducing \textit{Adaptive Chunking}, a framework that selects the most suitable chunking strategy for each document based on a set of five novel intrinsic, document-based metrics : \textit{References Completeness} (RC), \textit{Intrachunk Cohesion} (ICC), \textit{Document Contextual Coherence} (DCC), \textit{Block Integrity} (BI), and \textit{Size Compliance} (SC), which directly assess chunking quality accross key dimensions. To support this framework, we also introduce two new chunkers, an \textit{LLM-regex splitter} and a \textit{split-then-merge recursive splitter}, alongside targeted post-processing techniques. On a diverse corpus spanning legal, technical, and social science domains, our metric-guided adaptive method significantly improves downstream RAG performance. Without changing models or prompts, our framework increases RAG outcomes, raising answers correctness to 72\% (from 62-64\%) and increasing the number of successfully answered questions by over 30\% (65 vs. 49). These results demonstrate that adaptive, document-aware chunking, guided by a complementary suite of intrinsic metrics, offers a practical and effective path to more robust RAG systems. Code available at \url{https://github.com/ekimetrics/adaptive-chunking}.
\\ \newline }

\hypersetup{
  pdftitle={Adaptive Chunking: Optimizing Chunking-Method Selection for RAG},
  pdfauthor={Paulo Roberto de Moura Júnior, Jean Lelong, Annabelle Blangero}
}

\begin{document}

\maketitleabstract

\section{Introduction and Related Work}
Retrieval-Augmented Generation (RAG) has become a key paradigm for enhancing Large Language Models (LLMs) by grounding them in external knowledge sources \citep{zhao2025mocmixturestextchunking, merola2025reconstructingcontextevaluatingadvanced, wang-etal-2025-document}.  In essence, RAG combines two steps: retrieval, which fetches relevant information from a knowledge base, and generation, where an LLM uses this retrieved context to produce more accurate and informed responses. The effectiveness of RAG strongly depends on the retrieval pipeline's ability to fetch relevant context from a knowledge base for each user query \citep{wang-etal-2025-document}. Despite the growing size of LLM context windows, document segmentation remains critical for efficient retrieval \citep{Smith-Troynikov-2024-Chroma-Chunking, IBM-Enhancing-RAG-Chunking-2025}. Current chunking approaches often impose arbitrary boundaries that break contextual links, a problem referred to as the context-preservation dilemma \citep{qu2024semanticchunkingworthcomputational, duarte2024lumberchunker,jain-etal-2025-autochunker}. This fragmentation scatters essential information across disconnected chunks, degrading retrieval quality \citep{duarte2024lumberchunker, Anthropic-Contextual-Retrieval-2024, merola2025reconstructingcontextevaluatingadvanced}.

Some existing chuncking methods attempt to maintain a degree of semantic coherence, but often at the expense of other essential properties. \textit{Sentence-based splitters} \citep{Qi-etal-2020-Stanza} divide text into a fixed numbers of semantically cohesive sentences, but often break logical blocks such as paragraphs. Recursive splitters, such as LangChain's \texttt{RecursiveCharacterTextSplitter} \citep{LangChain-RecursiveCharacterTextSplitter} apply a hierarchy of separators to enforce length constraints, but they can group unrelated topics together, reducing cohesion. \textit{Semantic chunkers} \citep{LangChain-SemanticChunker, ni2025crossformer, llamaindex_semantic_chunking} use embeddings to split text at semantic boundaries, improving cohesion but introducing high computational cost. Finally LLM-driven methods \citep{duarte2024lumberchunker, jain-etal-2025-autochunker}, leverage language models to infer natural boundaries, which improves completeness but still assumes a single uniform strategy for an entire corpus and incurs significant overhead.

Each of these methods comes with inherent drawbacks, a trade-off between properties, and they all share a common limitation: relying on one fixed chunking strategy for all documents, regardless of structural or contextual differences.

Recent studies increasingly recognize that a "one-size-fits-all" chunking strategy is inadequate \citep{zhao2025mocmixturestextchunking} and that the optimal approach is inherently dependent on the document's structure and content \citep{IBM-Enhancing-RAG-Chunking-2025}. A critical gap in evaluation exacerbates this reliance on a monolithic strategy, as most studies measure chunking quality extrinsically via downstream retrieval metrics such as Hits@k, Recall@K, and Normalized Discounted Cumulative Gain (nDCG@K) \citep{günther2025latechunkingcontextualchunk, Anthropic-Contextual-Retrieval-2024, merola2025reconstructingcontextevaluatingadvanced}, making it difficult to isolate the impact of the chunking strategy itself \citep{Smith-Troynikov-2024-Chroma-Chunking}. This highlights a pressing need for robust, intrinsic metrics capable of directly assessing chunk quality.

To optimize retrieval, we consider the hypothesis that ideal chunks should simultaneously satisfy four key properties. They should be (1) \textbf{self-contained and logically complete}, expressing a full thought that can be understood in isolation; (2) \textbf{respectful of length constraints}, balancing the need for sufficient context with the token limits of embedding models; (3) \textbf{semantically cohesive}, focusing on a single topic to avoid informational noise; and (4) \textbf{context-preserving}, aligning with the document's natural structure, such as paragraphs or sections. Each property addresses a failure mode identified in prior work: breaking logical units scatters essential information across chunks \citep{duarte2024lumberchunker, Anthropic-Contextual-Retrieval-2024}, oversized or undersized chunks degrade embedding quality and waste retrieval slots \citep{günther2025latechunkingcontextualchunk, Smith-Troynikov-2024-Chroma-Chunking}, and mixing unrelated topics dilutes semantic signals \citep{qu2024semanticchunkingworthcomputational}.

In this work, we propose \textbf{Adaptive Chunking}, a framework that moves beyond the single-strategy paradigm by selecting the most suitable chunking method for each document. Our contribution is threefold:

\begin{enumerate}
    \item A suite of five intrinsic metrics: References Completeness, Intrachunk Cohesion, Document Contextual Coherence, Block Integrity, and Size Compliance—that directly assess chunk quality;
    \item New chunking techniques, including an LLM-guided regex splitter and a split-then-merge recursive splitter, complemented by targeted post-processing to enforce size constraints;
    \item An evaluation pipeline that measures both retriever quality and downstream RAG performance using a novel contextual completeness metric.
\end{enumerate}
Together, these components enable document-aware chunking that improves retrieval and question answering without modifying models or prompts.

\section{Method}
\subsection{Document Collection and Pre-processing}

Our experiments use a real-world corpus of 33 PDF documents from the Ekimetrics CLAIR project, spanning the legal, technical, and social science domains. This collection is intentionally diverse in its formatting, vocabulary and length, making it an robust test case for evaluating optimized chunking strategies.

The documents were parsed into markdown text using Microsoft Azure AI Document Intelligence (ADI) \citep{Microsoft-ADI}, which provides high-quality structural outputs. We then implemented a custom Markdown generation pipeline based on ADI's section tree, extending its built-in capabilities with three enhancements:
\begin{itemize}
    \item Improved table handling by splitting oversized tables (>1000 tokens) into sub-tables, while preserving  headers and reducing unnecessary token overhead.
    \item grouping page headers and footers into single blocks to avoid redundant markers;
    \item preserving inseparable spans such as tables, figures, titles with body text, sequences of short items (<100 tokens), and footnotes with their preceding text.
\end{itemize}

After parsing and Markdown formatting, the total corpus comprises approximately 1.18 million tokens.

The overall statistics of the collection are presented in Table \ref{tab:corpus-stats}.

\begin{table}[t]
\centering
\setlength{\tabcolsep}{4pt}
\renewcommand{\arraystretch}{1.05}
\begin{tabular*}{\columnwidth}{@{\extracolsep{\fill}} l r r r @{}}
\hline
 & \textbf{Tech} & \textbf{Legal} & \textbf{Social} \\
\hline
Docs (count)          & 9       & 16      & 8       \\
Tokens / doc (mean)     & 5\,257  & 30\,895 & 79\,862 \\
Tokens / doc (max)      & 18\,605 & 118\,424& 391\,660 \\
Tokens / doc (min)      & 1\,475  & 523     & 8\,013  \\
Pages / doc (mean)    & 12      & 46      & 114     \\
\hline
Total corpus tokens  & 47\,313  & 494\,320 & 638\,896 \\
\hline
\end{tabular*}
\caption{Corpus statistics per domain. Tokens were counted using OpenAI \texttt{o200k\_base} tokenizer (Byte-pair encoding).}
\label{tab:corpus-stats}
\end{table}

\subsection{Chunking Methods}

We introduce two new chunking methods designed to balance structural awareness with efficiency:

\begin{enumerate}
    \item \textbf{LLM Regex splitter}: this method combines the structural analysis capabilities of a language model with the determinism of regex-based splitting. We prompt the LLM with chunking guidelines, an output schema, and an example, then provide a sample document segment (first 8,000 tokens in our experiments). The LLM generates a regex pattern that is applied to the full document using Python's regex library. This approach is particularly effective for structured texts such as legal documents, where respecting natural boundaries (e.g., article delimiters) is critical—even when articles vary in length or span multiple pages.

    \item \textbf{Split-then-Merge Recursive Splitter}:  Inspired by LangChain's recursive splitter \citep{LangChain-RecursiveCharacterTextSplitter}, our variant introduces a two-pass strategy to reduce tiny, context-poor chunks \citep{günther2025latechunkingcontextualchunk}. In the first pass, the text is recursively split using a prioritized separator list (titles → sections → sentences → characters) until each segment is $\leq S$, where $S$ is the chunk size. In the second pass, adjacent segments are greedily merged without exceeding $S$, backtracking when necessary to maintain overlap and re-splitting oversized parts. This design improves size compliance and preserves context compared to single-step recursive methods. In our experiments, we include two variants of our recursive splitter (target sizes: 1,100 and 600 tokens).
    \end{enumerate}

Both methods integrate seamlessly with our Markdown pipeline, leveraging structural cues for better alignment with document hierarchy.
\\
\\
As final regularization steps, we propose two post-processing passes.

\begin{enumerate}
    \item \textbf{Oversized-chunk splitting}: Chunks exceeding the maximum size (1,100 tokens in our experiments) are re-split using the same separator cascade, leaving well-sized chunks untouched. This prevents embedding vectors from representing too many distinct ideas, which can dilute semantic signals \citep{günther2025latechunkingcontextualchunk} and obscure key information during retrieval \citep{wang-etal-2025-document}.

    \item \textbf{Tiny-chunk merging}: Chunks smaller than the minimum size (100 tokens) are merged with adjacent segments, provided the combined size remains below the upper limit (1,150 tokens). This reduces context-poor fragments that waste retrieval slots and degrade overall context quality.
    These steps ensure size compliance without compromising document structure.
\end{enumerate}

For comparison, we also evaluate baseline strategies:
\begin{itemize}
    \item Page-based chunking, i.e. chunking the documents by pages (with and without post-processing).
    \item LangChain's recursive splitter (with default parameters and with the same parameters as our custom recursive splitter).
    \item LangChain's experimental semantic splitter \citep{LangChain-SemanticChunker} with \texttt{gradient} thresholding.
    \item Sentence-based chunking, i.e. chunking by a fixed number of sentences (in our experiments, 5 sentences), using Stanza NLP model \citep{Qi-etal-2020-Stanza}.
\end{itemize}
Baseline methods are evaluated without post-processing, to preserve how they were designed and how they are typically used in practice.

\begin{table*}[t]
\centering
\setlength{\tabcolsep}{2pt}
\renewcommand{\arraystretch}{1.05}
\begin{tabular}{%
  >{\raggedright\arraybackslash}p{4.5cm}%
  *{6}{>{\centering\arraybackslash}p{1.75cm}}%
}
\hline
\textbf{Chunking method} & \textbf{mean} & \textbf{max}& \textbf{min}& \textbf{std. dev.} & \textbf{\# chunks}& \textbf{time [$s$]} \\
\hline
${}^{*}$ LLM regex (GPT-5)                 & $518$ & $1146$ & $69$  & $332$  & $2279$ & \textbf{${}^{**}146.85$} \\
${}^{*}$ our recursive ($s=1100$)       & $878$ & $1141$ & $104$ & $217$  & $1345$ & $31.42$ \\
${}^{*}$ our recursive ($s=600$)        & $496$ & $691$  & $102$ & $101$  & $2381$ & $28.3$ \\
${}^{*}$ page (post-processed)             & $663$ & $1146$ & $72$  & $235$  & $1780$ & $4.09$ \\
\hline
${}^{\dagger}$ LC recursive ($s=1100$)        & $706$ & $1101$ & $2$   & $324$  & $1675$ & $5.63$ \\
${}^{\dagger}$ LC recursive (default)         & $773$ & $1364$ & $57$  & $129$  & $1557$ & $0.04$ \\
${}^{\dagger}$ page (raw)                     & $669$ & $1765$ & $0$   & $277$  & $1765$ & $0.01$ \\
${}^{\dagger}$ semantic                       & $693$ & $17146$ & $1$  & $1095$ & $1706$ & $389.25$ \\
${}^{\dagger}$ sentence                       & $146$ & $1025$ & $4$   & $68$   & $8125$ & $30.43$ \\
\hline
Adaptive Chunking                             & $724$ & $1146$ & $86$  & $247$  & $1631$ & $210.66$ \\
\hline
\end{tabular}
\caption{Chunk sizes in tokens (OpenAI's \texttt{o200k\_base}) and runtime statistics. Chunking methods marked with ${}^{*}$ are included in the Adaptive Chunking results. Those marked with ${}^{\dagger}$ were not post-processed; ** was computed considering asynchronous API calls.}
\label{tab:metametrics}
\end{table*}

\subsection{Chunking Metrics}
We introduce five intrinsic metrics to evaluate chunk quality at the document level.
All metrics are computed at the document level, enabling both per-document tuning and corpus-level optimization.

\subsubsection{References Completeness (RC)}

Measures the fraction of entity–pronoun pairs that remain intact within a chunk. Breaking these pairs across boundaries can lead to incomplete retrieval when queries reference pronouns without their antecedents.
The metric computation requires, first, the extraction of mention clusters using Maverick coreference resolution model \citep{Martinelli-etal-2024-Maverick}, and then we extract the entity-pronoun pairs from the clusters, mapping pronouns to entities. Let $P = \{(e_i,p_i)\}_{i=1}^{N}$ denote the set of entity-pronoun pairs, with $s_i = \mathrm{start}(e_i)$ and $t_i = \mathrm{end}(p_i)$ their respective span boundaries. Let $B$ be the set of interior chunk boundaries, i.e., all chunk starts except the very first and all chunk ends except the very last. \textbf{The Reference Completeness (RC)} score is defined as
\begin{gather*}
m_i = \mathbf{1}\left[\exists\, b\in B:\; s_i < b \leq t_i\right],\\
\mathrm{RC} = 1 - \frac{1}{N}\sum_{i=1}^{N} m_i,
\end{gather*}
where $m_i$ is an indicator that equals $1$ if any boundary lies strictly between the entity and its pronoun, in which case the reference is considered missing.
Note that \textbf{RC} can only be computed for English-language documents, as the Maverick coreference resolution model is limited to English text.

\begin{figure}[t]
\centering
\begin{tikzpicture}
\pgfplotstableread[row sep=\\,col sep=&]{
x & y & z \\
1 & 1 & 1.00\\ 2 & 1 & -0.21\\ 3 & 1 & 0.24\\ 4 & 1 & 0.31\\ 5 & 1 & 0.04\\
1 & 2 & -0.21\\ 2 & 2 & 1.00\\ 3 & 2 & -0.44\\ 4 & 2 & -0.34\\ 5 & 2 & 0.12\\
1 & 3 & 0.24\\ 2 & 3 & -0.44\\ 3 & 3 & 1.00\\ 4 & 3 & 0.31\\ 5 & 3 & -0.16\\
1 & 4 & 0.31\\ 2 & 4 & -0.34\\ 3 & 4 & 0.31\\ 4 & 4 & 1.00\\ 5 & 4 & 0.02\\
1 & 5 & 0.04\\ 2 & 5 & 0.12\\ 3 & 5 & -0.16\\ 4 & 5 & 0.02\\ 5 & 5 & 1.00\\
}\corrtable
\pgfplotstableread[row sep=\\,col sep=&]{
x & y & z \\
1 & 2 & -0.21\\
1 & 3 & 0.24\\ 2 & 3 & -0.44\\
1 & 4 & 0.31\\ 2 & 4 & -0.34\\ 3 & 4 & 0.31\\
1 & 5 & 0.04\\ 2 & 5 & 0.12\\ 3 & 5 & -0.16\\ 4 & 5 & 0.02\\
}\corrlower

\begin{axis}[
  width=.78\columnwidth, height=.78\columnwidth,
  scale only axis, view={0}{90}, axis on top, y dir=reverse,
  xmin=0.5, xmax=5.5, ymin=0.5, ymax=5.5,
  enlargelimits=false, clip=false,
  tick style={draw=none},
  tick label style={font=\scriptsize},
  xtick={1,2,3,4,5}, xticklabels={RC, ICC, DCC, BI, SC},
  ytick={1,2,3,4,5}, yticklabels={RC, ICC, DCC, BI, SC},
  point meta min=-1, point meta max=1,
  colorbar, colormap/viridis,
  colorbar style={width=1.5mm, tick label style={font=\scriptsize}},
]
  \addplot[
    matrix plot*, mesh/rows=5, point meta=explicit,
  ] table[x=x, y=y, meta=z] {\corrtable};
  \fill[white] (axis cs:1.5,0.5) rectangle (axis cs:5.5,1.5);
  \fill[white] (axis cs:2.5,1.5) rectangle (axis cs:5.5,2.5);
  \fill[white] (axis cs:3.5,2.5) rectangle (axis cs:5.5,3.5);
  \fill[white] (axis cs:4.5,3.5) rectangle (axis cs:5.5,4.5);
  \addplot[only marks, mark=none, point meta=explicit,
    nodes near coords={\pgfmathprintnumber[fixed,precision=2]{\pgfplotspointmeta}},
    nodes near coords style={font=\scriptsize, anchor=center, inner sep=0pt},
  ] table[x=x, y=y, meta=z] {\corrlower};
\end{axis}
\end{tikzpicture}
\caption{Spearman correlation (\(\rho\)) between chunking metrics, computed using the metric results for each document and chunking method.}
\label{fig:spearman-heatmap}
\end{figure}
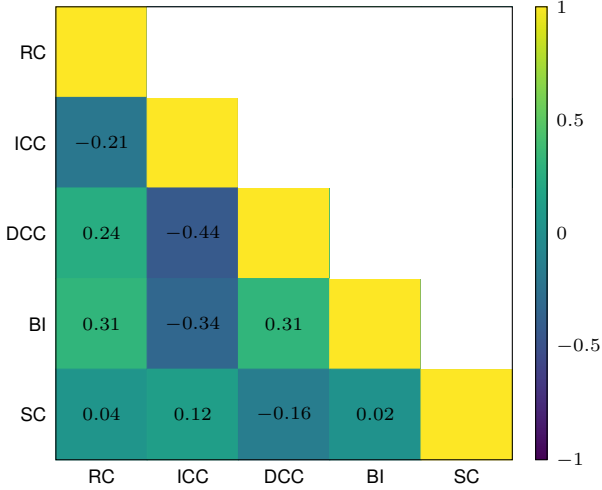

\subsubsection{Block Integrity (BI)}
Evaluates whether structural units, such as paragraphs, tables, figures, and title-body pairs, remain unbroken. Splitting these blocks reduces interpretability (e.g., a table split across chunks becomes unusable). \textbf{BI} uses parser-provided block spans and computes the proportion of intact blocks, applying a tolerance margin of 5 characters to avoid false positives.
Let the list of gold block boundaries be $G =\{0, d_1, \dots, d_M, L\}$ with $L$ the document length, and $B$ be the set of interior chunk boundaries. A block $[d_j, d_{j+1}]$ is considered broken if there exists a predicted chunk boundary $b \in B$ such that
\begin{eqnarray*}
    d_j + \tau < b < d_{j+1} - \tau,
\end{eqnarray*}
where $\tau$ is a tolerance margin in characters (in our experiments, $\tau=5$). Equivalently, the block is \emph{intact} if no such $b$ lies strictly inside it.  The Block Integrity score is the fraction of gold blocks that remain intact,
\begin{gather*}
I_j = \mathbf{1}\!\left[\not\exists\, b \in B:\; d_j + \tau < b < d_{j+1} - \tau \right],\\
\mathrm{BI} = \frac{1}{|G|-1} \sum_{j=0}^{|G|-1} I_j \, .
\end{gather*}

\subsubsection{Intrachunk Cohesion (ICC)}
Quantifies semantic consistency within a chunk by comparing sentence embeddings to the chunk embedding. For each chunk, ICC is the mean cosine similarity between its sentences and the full chunk embedding, computed using Jina AI's jina-embeddings-v3 model \citep{sturua2024jinaembeddingsv3multilingualembeddingstask}.
In more details, let the document be split into chunks $C = \{c_1, \dots, c_K\}$, and let each chunk $c_k$ contain a sequence of text blocks $S_k = \{s_{k1}, \dots, s_{kn_k}\}$ obtained using the starting indexes of text spans from the parsing outputs. Let $\mathbf{v}(c_k) \in \mathbb{R}^d$ denote the normalized embedding of chunk $c_k$, and $\mathbf{v}(s_{kj})$ the normalized embedding of sentence $s_{kj}$. The semantic cohesion of a single chunk is defined as the mean cosine similarity between its sentences and its full embedding,
\begin{eqnarray*}
\mathrm{Cohesion}(c_k) \;=\; \frac{1}{n_k} \sum_{j=1}^{n_k}
    \mathbf{v}(s_{kj})^\top \mathbf{v}(c_k),
\qquad n_k \ge 2,
\end{eqnarray*}
where chunks with fewer than two blocks are ignored, since cohesion cannot be meaningfully computed. The overall Intrachunk Cohesion (ICC) score is the mean across all valid chunks,
\begin{eqnarray*}
\mathrm{ICC} \;=\; \max\{0, \frac{1}{|\mathcal{K}|}\sum_{k\in\mathcal{K}} \mathrm{Cohesion}(c_k)\},
\end{eqnarray*}
where $\mathcal{K} \subseteq \{1,\dots,K\}$ indexes the chunks with
$n_k \ge 2$ and any negative score is clipped to zero.

\subsubsection{Document Contextual Coherence (DCC)}
Assesses how well chunks align with their broader local context \citep{günther2025latechunkingcontextualchunk}. We construct sliding windows of up to 3,000 tokens and compute the mean cosine similarity between each chunk and its window embedding (computed using Jina AI's jina-embeddings-v3 model \citep{sturua2024jinaembeddingsv3multilingualembeddingstask}). This ensures chunks are understandable in isolation while preserving document-level context.
As previously, let the document be segmented into ordered chunks $C=\{c_1,\dots,c_K\}$ with character spans $[a_k,b_k)$. Let $\mathrm{tok}(\cdot)$ denote a token-count function and let $T$ be a token budget (in our experiments, $T=3000$ tokens). We build a sequence of context windows $\{W_m\}$ as follows: starting at chunk index $i$, we concatenate the non-overlapping tails of consecutive chunks $c_i,c_{i+1},\dots$ (i.e., only the text not already included in the window) until the cumulative token count of the added tails would exceed $T$, always requiring each window to contain at least two chunks. The window starting index then advances by a stride of \texttt{window\_step} chunks, and the process repeats. This yields windows $W_m$ and their associated chunk index sets $\mathcal{C}_m \subset \{1,\dots,K\}$. Let $\mathbf{v}(c_k)\in \mathbb{R}^d$ be the \emph{normalized} embedding of chunk $c_k$, and let $\mathbf{w}(W_m)\in\mathbb{R}^d$ be the \emph{normalized} embedding of window $W_m$. For each window, define the window coherence via mean cosine similarity,
\begin{eqnarray*}
\mathrm{Coherence}(W_m) \;=\; \frac{1}{|\mathcal{C}_m|}\sum_{k \in \mathcal{C}_m }^{}{\mathbf{w}(W_m)^\top \mathbf{v}(c_k)},
\end{eqnarray*}
and finally aggregate all window coherence scores to build the Document Contextual Coherence as
\begin{eqnarray*}
\mathrm{DCC}  = \max\{ {0, \frac{1}{|M|}}\sum_{m=0}^{M} \mathrm{Coherence}(W_m) \},
\end{eqnarray*}
where $M$ is the total number of windows.

\begin{figure*}[t]
  \centering
  \includegraphics[width=0.8\textwidth]{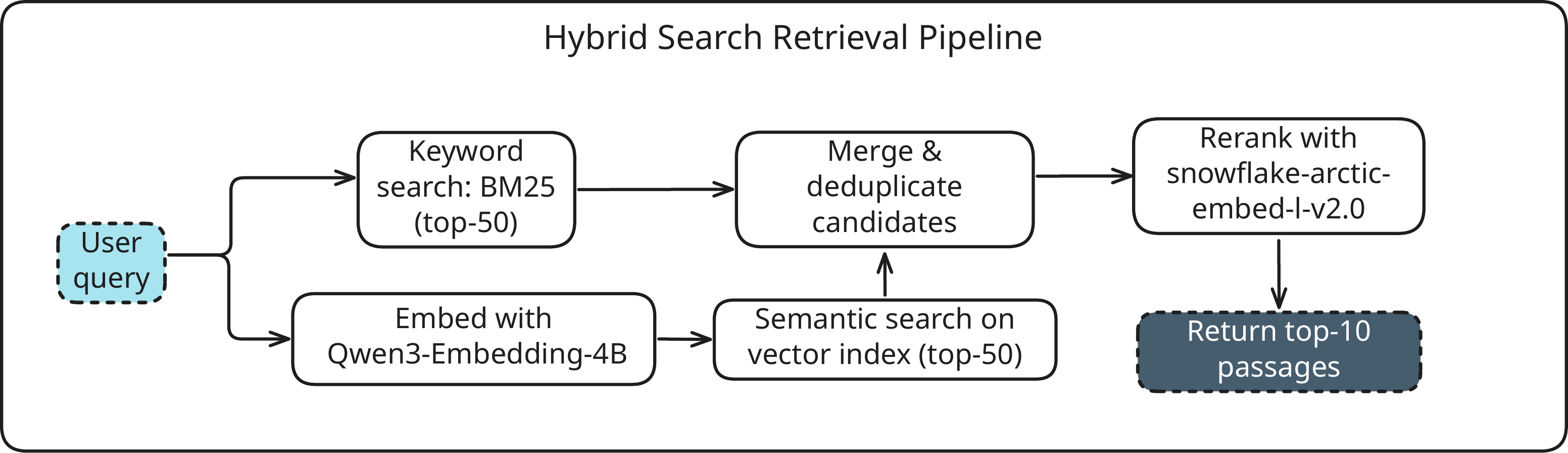}
  \caption{Hybrid search retrieval pipeline.}
  \label{fig:hybrid-search}
\end{figure*}

\subsubsection{Size Compliance (SC)}
Measures the proportion of chunks whose token length falls within predefined bounds (100–1,100 tokens in our experiments). Oversized chunks dilute embeddings, while tiny chunks waste retrieval slots and degrade context quality.
Let $C=\{ c_1,\dots,c_K\}$ be the set of chunks, $\mathrm{tok}(\cdot)$ a
token-count function, and $m$ and $M$ the minimum and maximum allowed tokens (in our experiments $m=100$, $M=1100$). For each chunk, let $\tau_k=\mathrm{tok}(c_k)$, then the score is the mean compliance
\begin{eqnarray*}
\mathrm{SC} \;=\; \frac{1}{K}\sum_{k=1}^{K} \mathbf{1}\big[\, m \le \tau_k \le M \,\big].
\end{eqnarray*}

\subsection{Retrieval and Experiments Setup}

\subsubsection{Pipeline Steps}
To evaluate the impact of chunking strategies on RAG performance, we implement a hybrid-search pipeline combining keyword-based and semantic retrieval, reranking, and generation (Figure \ref{fig:hybrid-search}). Here are the different steps:

\begin{enumerate}
    \item \textbf{Query Embedding}: Encode user queries using Qwen3-Embedding-4B, selected for strong performance on retrieval tasks.
    \item \textbf{Keyword Search}: Apply BM25 to retrieve the top-50 candidates based on lexical similarity.
    \item \textbf{Semantic Search}: Perform dense retrieval using cosine similarity between query embeddings and chunk embeddings, selecting the top-50 candidates.
    \item \textbf{Candidate Merging}: Combine BM25 and semantic results (100 candidates), deduplicate, and pass to reranking.
    \item \textbf{Reranking}: Use snowflake-arcticembed-l-v2.0, a state-of-the-art reranker optimized for hybrid retrieval, to select the top-10 most relevant chunks.
    \item \textbf{Answer Generation}: Provide these top-10 chunks as context to GPT-4.1 with temperature=0 and top-p=1 to minimize variability. The model is instructed to answer "I don't know based on the provided context" if the retrieved chunks are insufficient.
\end{enumerate}

All models, hyperparameters, and prompts remain fixed; only the chunking method varies to isolate its effect.

\subsubsection{Evaluation Setup}
To rigorously assess the impact of chunking strategies on RAG performance, we designed an evaluation protocol that compares three chunking strategies using two complementary metrics: one measuring retrieval quality and the other assessing the correctness of generated answers.
\\\\
\textbf{Chunking Strategies Compared}
\\
We benchmark three approaches:
\begin{enumerate}
    \item \textbf{Adaptive Chunking} which dynamically selects the best chunking method for each document based on the average of five intrinsic metrics.
    \item \textbf{LangChain Recursive Splitter} \citep{LangChain-RecursiveCharacterTextSplitter} with default parameters and no post-processing, representing a widely used baseline in retrieval systems.
    \item \textbf{Page-based Chunking} without post-processing, serving as the simplest and most common baseline in production pipelines.
\end{enumerate}

\textbf{Question–Answer Generation}
For each document in our corpus, we generate three question–answer pairs using GPT-4.1 with a 10,000-token context window. This ensures that questions are grounded in the document content and that answers reflect realistic retrieval scenarios.
\\\\
\textbf{Evaluation Metrics}
\begin{itemize}
    \item \textbf{Retrieval Completeness:} Based on DeepEval \citep{DeepEval-ConfidentAI-2025}, we designed this metric to measures how well the retrieved context supports the ground truth answer. GPT-4.1 acts as a judge, labeling each context as incomplete ($0$), partially complete ($1$), or complete ($2$). Scores are normalized to [0,1]. In our qualitative experiments, we observed that DeepEval's Contextual Recall produced inconsistent results for cases where the LLM had skipped answering queries due to insufficient context, whereas Retrieval Completeness yielded more consistent results.
    \item \textbf{Answer Correctness:} Using G-Eval implemented in DeepEval \citep{Liu-etal-2023-GEval, DeepEval-AnswerCorrectness-2025}, we compare the generated answer to the ground truth for factual alignment on a $0–1$ scale. This metric prioritizes correctness over stylistic differences and lightly penalizes omissions.
\end{itemize}

By combining these two metrics, we capture both retrieval quality and generation accuracy. This setup ensures that any observed performance differences can be attributed to chunking strategies rather than confounding factors such as model choice or prompt design.

\begin{table*}[t]
\centering
\setlength{\tabcolsep}{2pt}
\renewcommand{\arraystretch}{1.05}
\begin{tabular}{%
  >{\raggedright\arraybackslash}p{4.5cm}%
  *{6}{>{\centering\arraybackslash}p{1.75cm}}%
}
\hline
\textbf{Chunking method} & \textbf{RC} & \textbf{ICC} & \textbf{DCC} & \textbf{BI} & \textbf{SC} & \textbf{mean} \\
\hline
${}^{*}$ LLM regex (GPT-5)                & $98.0 \pm 2.9$  & $70.9 \pm 5.1$ & $82.4 \pm 5.5$ & $98.1 \pm 2.5$ & $99.6 \pm 1.3$ & $89.80$ \\
${}^{*}$ our recursive ($s=1100$)      & $\mathbf{99.0 \pm 1.8}$  & $66.6 \pm 3.8$ & $\mathbf{89.7 \pm 2.5}$ & $98.1 \pm 1.9$ & $\mathbf{100.0 \pm 0.1}$ & $\mathbf{90.68}$\\
${}^{*}$ our recursive ($s=600$)       & $97.2 \pm 2.9$  & $69.6 \pm 4.1$ & $84.7 \pm 3.1$ & $94.8 \pm 3.6$ & $\mathbf{100.0 \pm 0.0}$ & $89.24$ \\
${}^{*}$ page (post-processed)            & $97.2 \pm 3.5$  & $69.2 \pm 3.6$ & $86.4 \pm 4.6$ & $99.9 \pm 0.3$ & $99.9 \pm 0.4$ & $90.52$ \\
\hline
${}^{\dagger}$ LC recursive ($s=1100$)       & $98.4 \pm 3.1$  & $64.7 \pm 3.4$ & $86.8 \pm 2.8$ & $98.6 \pm 1.4$ & $93.3 \pm 7.2$ & $88.37$ \\
${}^{\dagger}$ LC recursive (default)        & $96.1 \pm 4.3$  & $65.6 \pm 3.6$ & $88.8 \pm 2.4$ & $95.0 \pm 2.8$ & $97.7 \pm 6.3$ & $88.62$ \\
${}^{\dagger}$ page (raw)                     & $97.1 \pm 3.5$  & $69.3 \pm 3.4$ & $86.1 \pm 5.0$ & $\mathbf{100.0 \pm 0.0}$ & $92.7 \pm 9.7$ & $89.03$ \\
${}^{\dagger}$ semantic                       & $97.5 \pm 3.1$  & $69.3 \pm 4.1$ & $76.3 \pm 7.3$ & $91.3 \pm 4.0$ & $48.1 \pm 17.6$ & $76.49$ \\
${}^{\dagger}$ sentence                       & $86.3 \pm 11.1$ & $\mathbf{78.4 \pm 2.7}$ & $72.5 \pm 5.8$ & $61.9 \pm 10.3$ & $67.2 \pm 23.1$ & $73.26$ \\
\hline
\textbf{Adaptive Chunking}                   & \textbf{$99.0 \pm 1.5$} & $68.2 \pm 4.7$ & $88.8 \pm 3.5$ & $99.4 \pm 1.2$ & $99.9 \pm 0.3$ & $\mathbf{91.07}$\\
\hline
\end{tabular}
\caption{Chunking performance results (\% mean $\pm$ \% st. dev.) for \textit{References Completeness} (RC), \textit{Intrachunk Cohesion} (ICC), \textit{Document Contextual Coherence} (DCC), \textit{Block Integrity} (BI), and \textit{Size Compliance} (SC). ``LC" denotes LangChain, and $s$ is the chunk size parameter in tokens. Rows marked with ${}^{*}$ are included in the Adaptive Chunking results; those with ${}^{\dagger}$ were not post-processed and are here for reference and comparison. Differences between Adaptive Chunking and all individual methods are statistically significant (Wilcoxon signed-rank test, $p < 0.001$).}
\label{tab:chunking-metrics}
\end{table*}

\section{Results}
\label{sec:results}
\subsection{Chunking Quality Analysis}
\textbf{Overview.}
\autoref{tab:chunking-metrics} reports our five intrinsic metrics: \textit{References Completeness} (RC), \textit{Intrachunk Cohesion} (ICC), \textit{Document Contextual Coherence} (DCC), \textit{Block Integrity} (BI), and \textit{Size Compliance} (SC), together with their mean across documents. The split-then-merge recursive splitter with a target size of $s{=}1100$ tokens obtains the highest overall mean (90.68\%), followed closely by post-processed page chunking (90.52\%) and the LLM-guided regex method (89.80\%). The Adaptive policy, which picks the best method per document, yields the highest corpus-level mean (91.07\%), indicating that no single method dominates across all documents.
\\
\smallskip\noindent
\textbf{Per-metric bests.} Method-level winners are consistent with the intended design trade-offs:
\begin{itemize}
  \item \textit{RC:} our recursive ($s{=}1100$) reaches the top score (99.0\%~$\pm$~1.8), suggesting fewer entity--pronoun splits.
  \item \textit{ICC:} sentence-based chunking is best (78.4\%~$\pm$~2.7), as smaller units reduce topical mixing.
  \item \textit{DCC:} our recursive ($s{=}1100$) achieves the highest coherence with local context (89.7\%~$\pm$~2.5).
  \item \textit{BI:} raw page chunking is perfect (100.0\%~$\pm$~0.0) because structural blocks rarely cross page boundaries.
  \item \textit{SC:} both recursive variants hit 100.0\% SC, confirming the effectiveness of split-then-merge size control.
\end{itemize}
The corresponding chunk sizes and runtime statistics are shown in \autoref{tab:metametrics}.

\subsection{Effect of Post-processing on Size Regularization}
\label{sec:results:postproc}
Post-processing has a clear and consistent effect. Methods with the tiny-chunk merge and oversized-chunk re-split maintain substantially higher \emph{minimum} chunk sizes (69--104 tokens) than their raw counterparts, which can produce near-empty fragments (0--4 tokens) (\autoref{tab:metametrics}). These small fragments carry little semantic content yet may intrude into the top-$k$ candidate set, displacing informative chunks. Empirically, post-processing improves SC and the mean intrinsic score by \textbf{+6 to +16} percentage points across methods, confirming the importance of enforcing size bounds without altering well-formed segments.

\subsection{Metric Complementarity and Trade-offs}
\label{sec:results:correlations}
Spearman correlations in \autoref{fig:spearman-heatmap} are weak to moderate ($-0.44{<}\rho{<}0.31$), implying that the five metrics capture complementary phenomena rather than a single latent factor. Two notable tensions emerge:
\begin{enumerate}
  \item \textbf{ICC vs.\ DCC.} Smaller chunks are more internally cohesive (higher ICC) but lose surrounding signal (lower DCC). Conversely, larger chunks preserve local context windows (higher DCC) but risk topical mixing (lower ICC).
  \item \textbf{ICC vs.\ BI.} Keeping structural units intact (high BI) may place heterogeneous elements (e.g., title + table + prose) in the same segment, which depresses ICC.
\end{enumerate}
These trade-offs justify a multi-metric objective and motivate a per-document selection policy rather than a corpus-wide ``best'' chunker.

\subsection{Adaptive Selection Behavior}
\label{sec:results:selection}
Although the recursive splitter with $s{=}1100$ achieves the highest average intrinsic score, \autoref{tab:chunking-selection} shows that the Adaptive policy chooses \emph{page (post-processed)} for \textbf{48\%} of documents and \emph{recursive $s{=}1100$} for \textbf{42\%}; the LLM-regex and recursive $s{=}600$ variants account for the remainder (6\% and 3\%). This dispersion reflects corpus heterogeneity: documents with strong pagination semantics (e.g., tables, footnotes, or article headers aligned to pages) benefit from page-level integrity, while long narrative or hierarchical content benefits from recursive segmentation tuned to size and overlap. The selection data substantiate the central claim that \emph{no single method is universally optimal}.

\begin{table}[t]
\centering
\setlength{\tabcolsep}{6pt}
\renewcommand{\arraystretch}{1.05}
\begin{tabular*}{\columnwidth}{@{\extracolsep{\fill}} l c @{}}
\hline
\textbf{Selected method} & \textbf{\% selection} \\
\hline
page (post-processed) & 48 \\
our recursive ($s = 1100$) & 42 \\
LLM regex (GPT-5) & 6  \\
our recursive ($s = 600$)  & 3  \\
\hline
\end{tabular*}
\caption{Results for chunking method selection in the Adaptive Chunking method with respect to the entire document corpus. Other chunking methods were not selected and omitted here.}
\label{tab:chunking-selection}
\end{table}

\subsection{Impact on Retrieval and Generation}
\label{sec:results:rag}
\textbf{Retrieval quality.} \autoref{tab:rag-results}
shows that Adaptive Chunking improves Retrieval Completeness to \textbf{67.68\%}, outpacing LangChain recursive default (58.08\%, \textbf{+9.60 pp}) and raw page chunking (59.09\%, \textbf{+8.59 pp}). In relative terms, these are \textasciitilde16.5--18.0\% gains. Since the retrieval configuration and models are held constant, these deltas are attributable to chunking alone.

\smallskip\noindent
\textbf{Answer correctness.} Adaptive Chunking also yields higher G-Eval correctness (\textbf{78.01\%}) than both baselines (70.11\%, \textbf{+7.90 pp} vs.\ LC recursive; 73.33\%, \textbf{+4.68 pp} vs.\ page). This shows that better retrieval context translates into more accurate answers under the same generator (GPT-4.1, $T{=}0$, $p{=}1$).

\smallskip\noindent
\textbf{Answer rate.} The system with Adaptive Chunking answers \textbf{65/99} queries versus \textbf{49/99} for each baseline (\textbf{+16} questions, \textbf{+32.7\%}). The increase is consistent with higher Retrieval Completeness: when the retrieved context is more often complete, the model is less likely to abstain (``I don't know based on the provided context.'').

\smallskip\noindent
\textbf{Amplification effect.} The intrinsic metric differences between the top chunking methods in \autoref{tab:chunking-metrics} are modest (0.4--2.4 percentage points), yet they translate into substantially larger RAG performance gaps (8--10~pp in Retrieval Completeness, 5--8~pp in Answer Correctness). This suggests that intrinsic chunk quality improvements compound through the retrieval and generation pipeline: slightly better chunks lead to more relevant retrieved passages, which in turn yield more complete and accurate answers.

\begin{table*}[t]
\centering
\setlength{\tabcolsep}{4pt}
\renewcommand{\arraystretch}{1.05}
\begin{tabular*}{\textwidth}{@{\extracolsep{\fill}} l c c c @{}}
\hline
\textbf{Metric} & \textbf{Adaptive Chunking} & \textbf{LC recursive (default)} & \textbf{page (raw)} \\
\hline
Retrieval Completeness & $\mathbf{67.68}$ & $58.08$ & $59.09$ \\
Answer Correctness     & $\mathbf{78.01}$ & $70.11$ & $73.33$ \\
Mean                   & $\mathbf{71.77}$           & $62.07$           & $63.80$ \\
\hline
Answered queries       & $\mathbf{65}$              & $49$              & $49$ \\
Total queries          & $99$                       & $99$              & $99$ \\
\hline
\end{tabular*}
\caption{RAG performance results (\% mean) evaluated using OpenAI's GPT-4.1 ($\textit{temperature}=0$) as the LLM judge. Retrieval Completeness is evaluated for all queries, while Answer Correctness skips queries where the model decided not to provide an answer (insufficient context). Highest scores per column are marked in bold; ``LC" means LangChain. Retrieval Completeness differences are statistically significant (Wilcoxon signed-rank, $p < 0.05$).}
\label{tab:rag-results}
\end{table*}

\subsection{Runtime and Scalability}
\label{sec:results:runtime}
\textbf{Chunking cost.} \autoref{tab:metametrics} indicates the LLM-regex approach has the highest average runtime (146.85s), driven by LLM calls (asynchronous). Our recursive splitters are efficient (28--31s) and scale well. The Adaptive pipeline's total (\textasciitilde210.66s) reflects running multiple candidate chunkers and computing metrics per document, which is amortized at indexing time.

\smallskip\noindent
\textbf{Evaluation cost.} \autoref{tab:eval-times} shows Document Contextual Coherence (15:58) and entity--pronoun extraction (13:13) dominate evaluation time, whereas computing chunk embeddings (3:04) and token metrics (0:17) are comparatively inexpensive. These bottlenecks suggest optimizations: (i) caching token-level embeddings once per document and composing window/chunk vectors; (ii) batched coreference clustering.

\subsection{Practical Implications}
\label{sec:results:implications}
The results support three practical recommendations for building RAG over heterogeneous corpora:
\begin{enumerate}
  \item \textbf{Prefer document-aware selection over a global default.} Even when a single chunker has the best average score, per-document selection delivers higher retrieval completeness and answer rates.
  \item \textbf{Always regularize sizes.} Tiny-chunk merging and oversized re-splitting provide consistent gains at negligible additional cost, improving both SC and the aggregate mean.
  \item \textbf{Balance cohesion and context.} Use multi-metric scoring (ICC+BI+DCC+RC+SC) rather than optimizing one dimension in isolation; this avoids overfitting to cohesion at the expense of context (or vice versa).
\end{enumerate}
\begin{table}[t]
\centering
\setlength{\tabcolsep}{5pt}
\renewcommand{\arraystretch}{1.05}
\begin{tabular*}{\columnwidth}{@{\extracolsep{\fill}} l c @{}}
\hline
\textbf{Step} & \textbf{Time (mm:ss)} \\
\hline
Document Contextual Coherence   & 15:58 \\
Intrachunk Cohesion             & 03:29 \\
Size Compliance                 & 00:07 \\
References Completeness         & 00:02 \\
Block Integrity                 & 00:01 \\
Entity-pronoun pairs extraction & 13:13 \\
Chunk embeddings                & 03:04 \\
Chunk token metrics             & 00:17 \\
\hline
\textbf{Total}                  & \textbf{36:11} \\
\hline
\end{tabular*}
\caption{Runtime per evaluation component. All embeddings were computed using \texttt{jina-embeddings-v3} sentence transformer, and the entity-pronoun pairs were extracted using Sapienza NLP \texttt{maverick-mes-ontonotes}.}
\label{tab:eval-times}
\end{table}

\section{Conclusion}
This work introduced \textit{Adaptive Chunking}, a framework for document-specific chunking in Retrieval-Augmented Generation (RAG). Our approach combines a robust parser-to-Markdown pipeline, two novel chunkers (LLM-guided regex and split-then-merge recursive splitter), and targeted post-processing to enforce size constraints. We proposed five intrinsic metrics: \textit{References Completeness} (RC), \textit{Intrachunk Cohesion} (ICC), \textit{Document Contextual Coherence} (DCC), \textit{Block Integrity} (BI), and \textit{Size Compliance} (SC), to enable document-level evaluation and guide chunker selection. Experiments on a heterogeneous corpus show that adaptive, metric-driven selection consistently improves retrieval completeness and answer correctness without modifying models or prompts. These findings demonstrate that principled, document-aware chunking offers a practical path to more robust RAG systems.

\section{Ethics Statement}
All experiments were conducted on publicly available documents from legal, technical, and social science domains. No personal or sensitive data were used. The proposed methods do not involve human subjects, demographic profiling, or any form of discriminatory processing. All evaluations were performed using automated systems under controlled conditions, ensuring compliance with ethical standards for reproducibility and transparency.

\section{Limitations}

\paragraph{Computational Efficiency}
Our evaluation pipeline introduces computational overhead (Table \ref{tab:eval-times}), primarily from Document Contextual Coherence and entity-pronoun extraction. DCC currently recomputes token embeddings for each sliding window; computing all token embeddings once at the document level and mapping them to windows would significantly reduce this cost. Entity-pronoun extraction is constrained by Maverick's lack of batch clustering support; a custom batched implementation could accelerate this step substantially.

\paragraph{Language and Domain Coverage}
The Maverick coreference model limits References Completeness to English text only, constraining multilingual applicability. Our corpus focuses on formal documents (legal, technical, social science); generalization to informal text, creative writing, or heavily multimodal content requires further validation. While we report detailed results on one corpus for reproducibility, we validated our approach on additional datasets with consistent performance gains.

\paragraph{Hyperparameters and Flexibility}
Hyperparameter choices (chunk size bounds of 100-1,100 tokens, sliding window size of 3,000 tokens, equal metric weighting) remain heuristic and user-dependent. Our metric calculations assume contiguous chunks matching the parsed Markdown text exactly, enabling precise character offset mapping but limiting evaluation of methods that assemble non-contiguous chunks or modify content. Block Integrity requires parser-generated block span annotations; users starting with pre-parsed text in a non-markdown format must provide equivalent annotations.

\paragraph{Practical Deployment}
Chunking strategies are selected once at indexing time and do not adapt to query characteristics or task requirements. Our evaluation uses a fixed hybrid search pipeline ; while this isolates the impact of chunking, performance with alternative embedding models, retrieval methods, or reranking strategies remains unexplored. Running multiple candidate chunkers per document may challenge high-throughput production pipelines, though this cost is amortized at indexing and could be reduced through pre-screening mechanisms.

\nocite{*}
\section{Bibliographical References}\label{sec:reference}

\bibliographystyle{lrec2026-natbib}
\bibliography{bibliography}

\appendix
\onecolumn

\renewcommand{\thesection}{Appendix \Alph{section}}
\renewcommand{\thesubsection}{\Alph{section}.\arabic{subsection}}

\section{Example of Generated Markdown Output}
\label{app:md-example}
\begin{lstlisting}[style=mdpretty]

# The Hamburg Commissioner for Data protection and freedom of information

## Discussion Paper: Large Language Models and Personal Data

This discussion paper reflects the current state of knowledge and understanding at the Hamburg Com- missioner for Data Protection and Freedom of Information (HmbBfDI) regarding the applicability of the General Data Protection Regulation (GDPR) to Large Language Models1 (LLMs). This paper aims to stim- ulate further debate. It is intended to support companies and public authorities in better navigating com- plex data protection issues surrounding this subject matter. To this end, this paper explains relevant tech- nical aspects of LLMs, assesses them in light of case law regarding personal data from the Court of Jus- tice of the European Union (CJEU) and highlights the resulting practical implications. From this, three principle theses can be derived:

1. The mere storage of an LLM does not constitute processing within the meaning of article 4 (2) GDPR. This is because no personal data is stored in LLMs. Insofar as personal data is processed in an LLM-supported AI system, the processing must comply with the require- ments of the GDPR. This applies in particular to the output of such an AI system.

2. Given that no personal data is stored in LLMs, data subject rights as defined in the GDPR cannot relate to the model itself. However, claims for access, erasure or rectification can certainly relate to the input and output of an AI system of the responsible provider or de- ployer.

3. The training of LLMs using personal data must comply with data protection regulations. Throughout this process, data subject rights must also be upheld. However, potential vio- lations during the LLMs training phase do not affect the lawfulness of using such a model within an AI system.

\* 1 This refers exclusively to models as an important, but not sole, component of a comprehensive AI system (e.g. an LLM-based chatbot).

<!-- PageFooter: www.datenschutz-hamburg.de E-mail: mailbox@datenschutz.hamburg.de Ludwig-Erhard-Strasse 22 - D-20459 Hamburg - Tel .: 040 - 4 28 54 - 40 40 - Fax: 040 - 4 28 54 - 40 00 Confidential information should only be sent to us electronically in encrypted form. Our public PGP key is available on the Internet (fingerprint: 0932 579B 33C1 8C21 6C9D E77D 08DD BAE4 3377 5707). -->
<!-- PageBreak -->

<!-- PageHeader: The Hamburg Commissioner for Data protection and freedom of information -->

### I. Introduction

When an LLM, functioning as a component of an AI system, processes2 prompts (so-called "in- ference"), the LLM's output may contain information relating to natural persons, especially if the prompt specifically asks for it. This raises the question of whether personal data is stored in an LLM.
\end{lstlisting}

\clearpage
\section{LLM Regex Prompt}
\label{app:llmregexprompt}

\begin{lstlisting}[style=prompt]
<Task>
Your task is to split a long document into self-contained and logically complete chunks to be used in a Retrieval Augmented Generation (RAG) system. Given a document text, choose the best **unique** regular-expression to be used as a *delimiter* to split it into small chunks using the Python `re` engine and the `re.split` function.
</Task>

<Output requirements>
You **must** return only the answer in this format:
    <regex>regex pattern here</regex>
</Output requirements>

<Splitting guidelines>
    - The regex pattern **must** be valid.
    - The chunks should be self-contained, logically complete and not too large.
    - Do not split paragraphs.
    - Do not split tables, marked between <Table> </Table> tags.
    - Do not split figures, marked between <Figure> </Figure> tags.
    - Do not split lists of short elements.
    - Do not split titles from the text that follows them.
    - Do not split footnotes from their parent text.
</Splitting guidelines>

<Splitting example>
    <Example of input text>
{example["input"]}
    </Example of input text>

    <Expected answer>
        <regex>{example["output"]}</regex>
    </Expected answer>
</Splitting example>

Now, please apply this method to the following text between <Input> and </Input> markers:
<Input>{document_context_str}</Input>
\end{lstlisting}

\clearpage
\section{Question-Answer Pairs Generation Prompt}
\label{app:qageneration-prompt}

\begin{lstlisting}[style=prompt]
<Task>
Create exactly {qa_pairs_per_document} high-quality, mutually distinct question-answer pairs based on the provided Document Lines context.
</Task>

<Instructions>
1. Create each question-answer pair based *only* on the text provided in the Document Lines section. Do not use any external knowledge.
2. Each question should be concise with less than 20 words and sound like a natural question a curious person would ask.
3. Each question should target a key concept or relationship, not a trivial detail.
4. Each question must be self-contained. They should still make sense if read alone. Avoid vague references (e.g., "the report/paper/it/they/this/that/above").
5. Answers must be strictly grounded in the document lines.
6. Write both questions and answers in English, even if the document lines are in another language.
7. Produce **{qa_pairs_per_document}** questions that are **meaningfully different** from one another: no rephrasing, duplicates, or overlapping focus. Cover different aspects of the content when possible.
</Instructions>

<Document Lines>
{document_context}
</Document Lines>
\end{lstlisting}

\section{Retrieval Completeness Evaluation Prompt}
\label{app:retrieval-completeness-prompt}

\begin{lstlisting}[style=prompt]
<Task>
You are an expert fact-checker. Your task is to evaluate how completely the provided Context supports a Reference Answer.
</Task>

<Instructions>
1.  Read the Context thoroughly.
2.  Read the Reference Answer carefully.
3.  Compare the Reference Answer against the Context.
4.  Classify the completeness level:
    - 0 = Incomplete: Context does not support key claims.
    - 1 = Partially Complete: Context supports some but not all claims.
    - 2 = Complete: Context fully supports all claims.
5.  Provide a brief one-sentence reason.
</Instructions>

<Context>
{context}
</Context>

<Reference Answer>
{reference_answer}
</Reference Answer>
\end{lstlisting}

\clearpage
\section{Answer Generation Prompt for RAG}
\label{app:answer-generation-prompt}

\begin{lstlisting}[style=prompt]
<Task>
You are a grounded QA assistant. Answer the question strictly using the provided context. Read all the documents provided as context before answering the question. Do not hallucinate. If the context is insufficient, reply exactly: "I don't know based on the provided context." and nothing else. Keep the answer concise. Answer only in English even if the context or question is in another language.
</Task>

<Context>
{context_str}
</Context>

<Question>
{question_str}
</Question>
\end{lstlisting}

\clearpage
\section{LLM Regex Splitter Diagram}
\label{app:llm-regex-splitter}

\begin{figure*}[h!]
  \centering
  \includegraphics[width=0.8\textwidth]{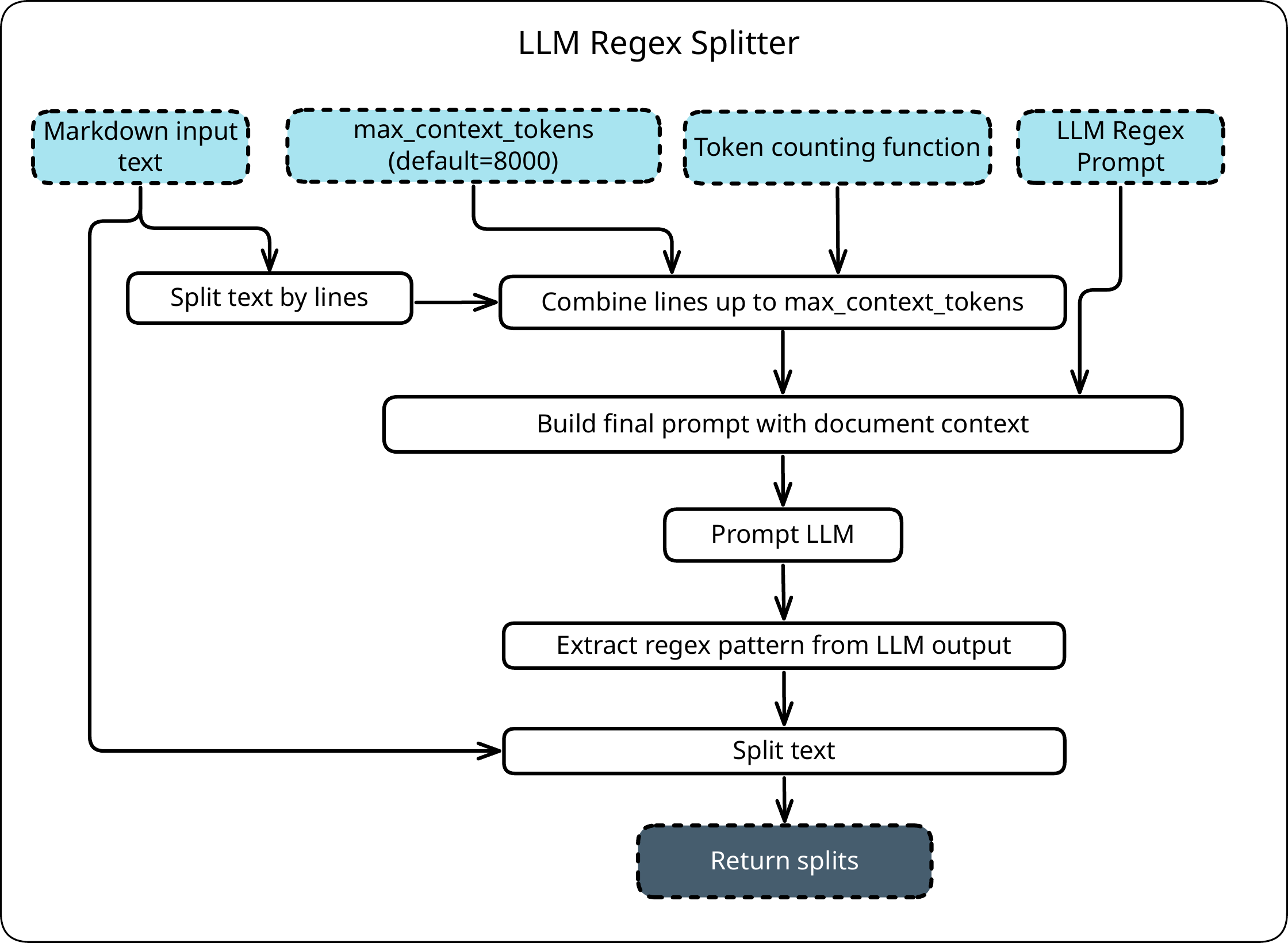}
  \caption{LLM Regex splitter pipeline. From markdown input and a token budget, the system builds a context-aware prompt, obtains a delimiter regex from the LLM, and applies it to split the document into chunks.}
\end{figure*}

\section{Separators list for recursive splitters}
\label{app:separators-list}

\begin{figure*}[h!]
  \centering
  \includegraphics[width=\textwidth]{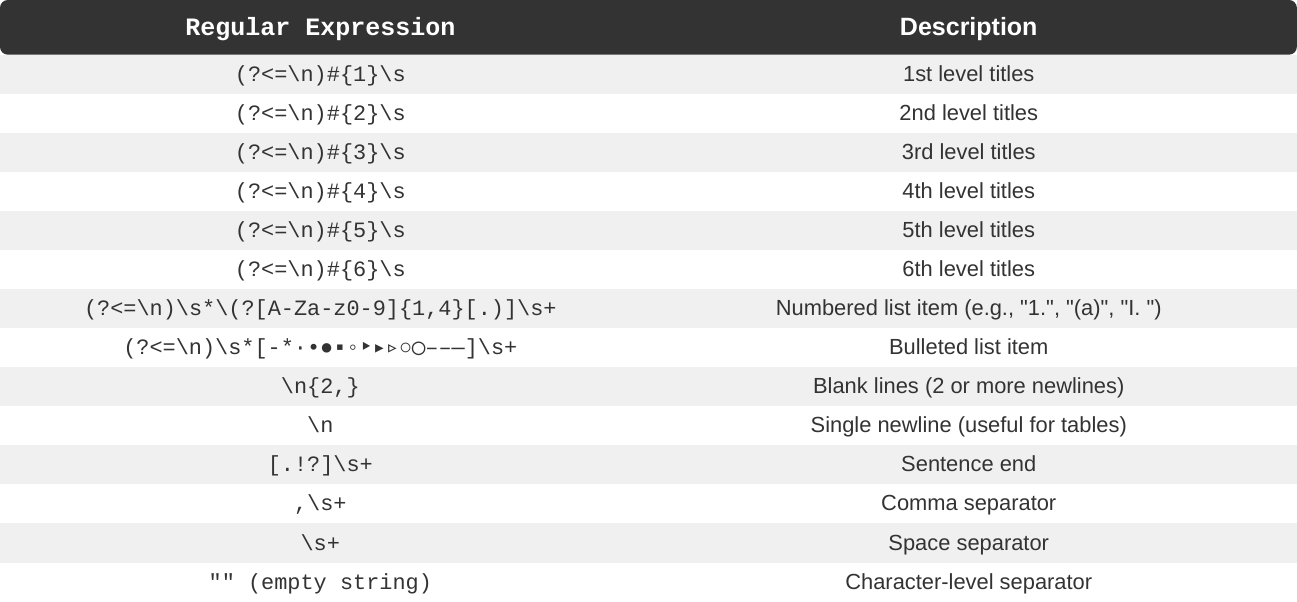}
  \caption{Regex separators list for recursive splitters, adapted to our Markdown parser outputs. The list is sorted from highest to lowest priority.}
\end{figure*}

\clearpage
\section{Split-then-merge Recursive Splitter Diagram}
\label{app:recursive-splitter}

\begin{figure*}[h!]
  \centering
  \includegraphics[width=\textwidth]{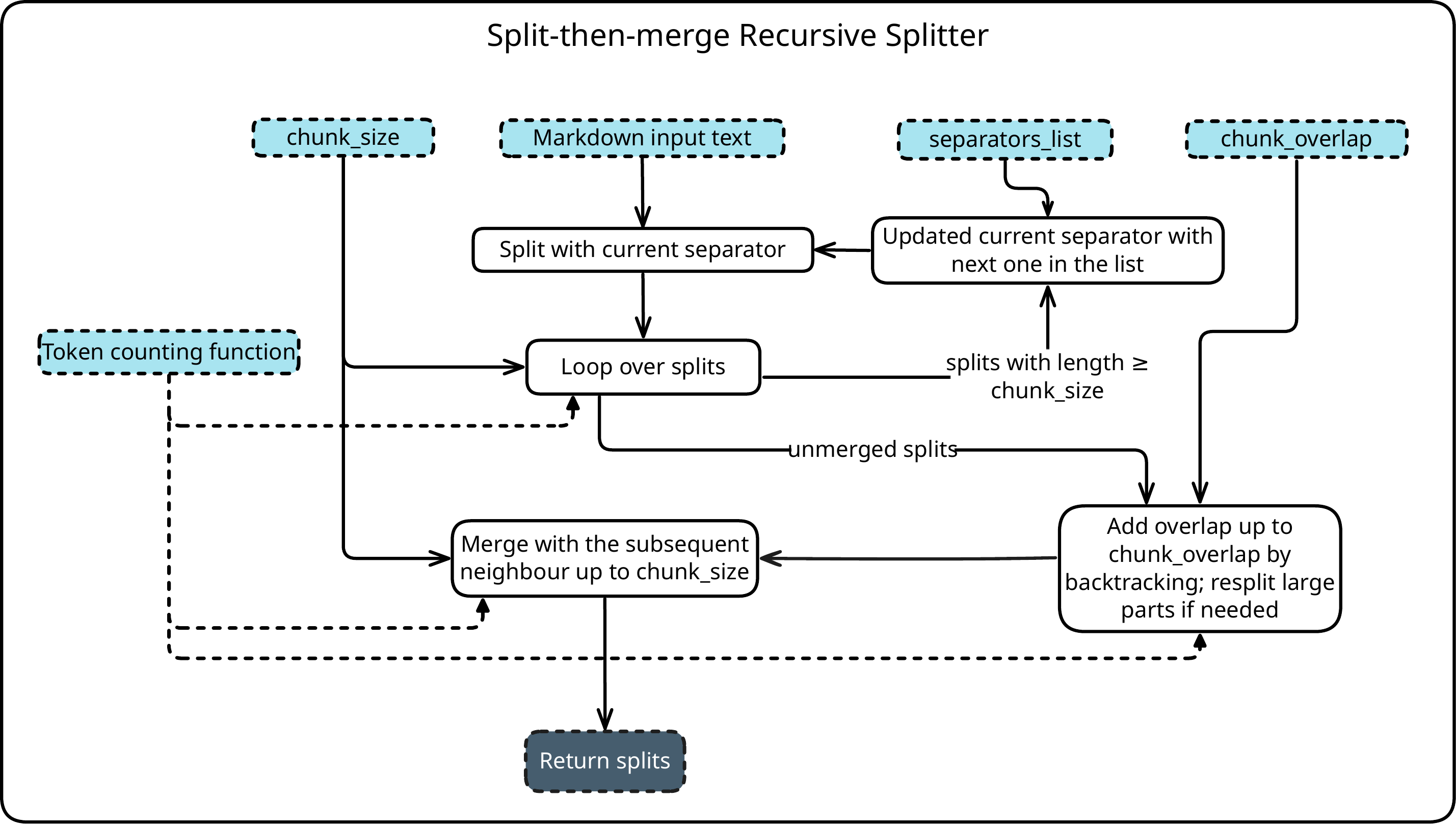}
  \caption{Split–then–merge recursive splitter pipeline. Text is recursively split following a priority list of separators until each piece is $\leq$  chunk size, then merged forward into chunks; when the next piece would exceed chunk size, a new chunk starts with up to chunk overlap tokens of backtracked overlap.}
\end{figure*}

\clearpage
\section{Post-Processing Effect Details}
\label{app:postprocessing}

Table \ref{tab:postprocessing-results-appendix} reports the detailed effect of our two-stage post-processing pipeline on Size Compliance (SC) and overall mean chunking scores. The pipeline consists of: (1) \emph{oversized re-split}, applied only to chunks exceeding 1100 tokens, and (2) \emph{tiny-chunk merge}, applied only to chunks smaller than 100 tokens, with a maximum merged size of 1150 tokens. The ``--'' symbol indicates that a particular step was not required for that method (e.g., our recursive splitters with $s=1100$ or $s=600$ already produce well-sized chunks in the raw output).

For every method that required post-processing, both SC and the final mean score improved substantially. The largest improvements were observed for the LLM regex method (raw SC: 58.3\% $\rightarrow$ final SC: 99.6\%, mean: 80.0\% $\rightarrow$ 89.8\%) and the semantic chunker (raw SC: 48.1\% $\rightarrow$ final SC: 99.9\%, mean: 76.5\% $\rightarrow$ 88.9\%), demonstrating that post-processing is particularly critical for methods that initially produce highly variable chunk sizes. The final SC scores do not always reach 100\% because we prioritize minimizing tiny chunks over strict adherence to the upper bound—specifically, we avoid merging fragments that would exceed 1150 tokens even if this leaves some chunks slightly below the 100-token minimum.

\begin{table*}[h!]
\centering
\setlength{\tabcolsep}{3pt}
\renewcommand{\arraystretch}{1.05}
\begin{tabular*}{\textwidth}{@{\extracolsep{\fill}} >{\raggedright\arraybackslash}p{5.0cm} cc cc cc @{}}
\hline
& \multicolumn{2}{c}{\textbf{Raw}} & \multicolumn{2}{c}{\textbf{Oversized re-split}} & \multicolumn{2}{c}{\textbf{Tiny-chunk merge}} \\
\textbf{Chunking method} & \textbf{SC \%} & \textbf{final mean \%} & \textbf{SC \%} & \textbf{final mean \%} & \textbf{SC \%} & \textbf{final mean \%} \\
\hline
LLM regex (GPT-5) & $58.3$  & $80.00$ & $71.4$  & $82.55$ & $99.6$  & $89.80$ \\
our recursive ($s =1100$)    & $100.0$ & $90.67$ & -- & -- & $100.0$ & $90.68$ \\
our recursive ($s=600$)     & $98.7$  & $88.96$ & --  & --  & $100.0$ & $89.24$ \\
page  & $92.7$  & $89.03$ & $95.3$  & $89.53$ & $99.9$  & $90.52$ \\
LC recursive ($s=1100$) & $93.3$  & $88.37$ & -- & -- & $99.4$  & $89.93$ \\
LC recursive (default)     & $97.7$  & $88.62$ & -- & -- & --  & -- \\
semantic  & $48.1$  & $76.49$ & $73.8$  & $82.11$ & $99.9$  & $88.90$ \\
sentence & $67.2$  & $73.26$ & -- & -- & $100.0$ & $82.03$ \\
\hline
\end{tabular*}
\caption{Effect of post-processing on Size Compliance (SC) and the final mean chunking score. Columns follow the sequential pipeline \emph{raw (no post-processing)} $\rightarrow$ \emph{oversized re-split} $\rightarrow$ \emph{tiny-chunk merge}. Oversized re-split is only applied to chunks with size $>1100$ tokens. Tiny-chunk merge is only applied to chunks with size $<100$ tokens with a maximum merging size of 1150 tokens.}
\label{tab:postprocessing-results-appendix}
\end{table*}

\end{document}